\documentclass[letterpaper]{article} 
\usepackage{aaai24}  
\usepackage{times} 
\usepackage{helvet}  
\usepackage{courier}  
\usepackage[hyphens]{url}  
\usepackage{graphicx} 
\usepackage{natbib}  
\usepackage{caption} 
\usepackage[linesnumbered]{algorithm2e}
\usepackage{xcolor}
\usepackage{booktabs}
\usepackage{multirow}
\usepackage{amsmath}
\usepackage{amsfonts}
\usepackage{mathrsfs}
\usepackage{siunitx}
\usepackage{colortbl}
\usepackage{xparse}
\usepackage{appendix}
\usepackage{glossaries}
\ExplSyntaxOn
\NewDocumentCommand{\longdash}{ O{2} }
 {
  --\prg_replicate:nn { #1 - 1 } { \negthinspace -- }
 }
\ExplSyntaxOff

\usepackage{newfloat}
\usepackage{hyperref}

\title{Reinforcement Learning as a Parsimonious Alternative to Prediction Cascades:\\ A Case Study on Image Segmentation}
\author{
    Bharat Srikishan\textsuperscript{\rm 1},
    Anika Tabassum\textsuperscript{\rm 2},
    Srikanth Allu\textsuperscript{\rm 2},\\
    Ramakrishnan Kannan\textsuperscript{\rm 2},
    Nikhil Muralidhar\textsuperscript{\rm 1}
}
\affiliations{
    \textsuperscript{\rm 1}Stevens Institute of Technology \\
    \textsuperscript{\rm 2}Oak Ridge National Laboratory \\
    bsrikish@stevens.edu, \{tabassuma, allus,
    kannanr\}@ornl.gov, nmurali1@stevens.edu
}

\newglossaryentry{ourmethod}
{
  name=PaSeR,
  description={Parsimonious Segmentation with Reinforcement Learning}
}
\newglossaryentry{ourmethodrandpolicy}
{
  name=PaSeR-RandPol.,
  description={Parsimonious Segmentation with random Reinforcement Learning Policy}
}
\newglossaryentry{idkbaseline}
{
  name=IDK-Cascade,
  description={I Don't Know Cascade}
}
\newglossaryentry{idkbaselineioumatch}
{
  name=IDK-Cascade (IoU Match),
  description={I Don't Know Cascade IoU Match}
}

\begin{document}

\newcommand{\pluseq}{\mathrel{{+}{=}}}
\newcommand{\bx}{\mathbf{x}}
\newcommand{\by}{\mathbf{y}}
\newcommand{\bz}{\mathbf{z}}
\newcommand{\be}{\mathbf{e}}
\newcommand{\bs}{\mathbf{s}}
\newcommand{\ba}{\mathbf{a}}
\newcommand{\btheta}{\mathbf{\theta}}
\newcommand{\hide}[1]{}

\newcommand\blfootnote[1]{
  \begingroup
  \renewcommand\thefootnote{}\footnote{#1}
  \addtocounter{footnote}{-1}
  \endgroup
}
\maketitle

\begin{abstract}
    Deep learning architectures have achieved state-of-the-art (SOTA) performance on computer vision tasks such as object detection and image segmentation. This may be attributed to the use of over-parameterized, monolithic deep learning architectures executed on large datasets. Although such large architectures lead to increased accuracy, this is usually accompanied by a larger increase in computation and memory requirements during inference. While this is a non-issue in traditional machine learning (ML) pipelines, the recent confluence of machine learning and fields like the Internet of Things (IoT) has rendered such large architectures infeasible for execution in low-resource settings. For some datasets, large monolithic pipelines may be overkill for simpler inputs. To address this problem, previous efforts have proposed \emph{decision cascades} where inputs are passed through models of increasing complexity until desired performance is achieved. However, we argue that cascaded prediction leads to sub-optimal throughput and increased computational cost due to wasteful intermediate computations. To address this, we propose PaSeR (Parsimonious Segmentation with Reinforcement Learning) a non-cascading, cost-aware learning pipeline as an efficient alternative to cascaded decision architectures. Through experimental evaluation on both real-world and standard datasets, we demonstrate that PaSeR achieves better accuracy while minimizing computational cost relative to cascaded models. Further, we introduce a new metric IoU/GigaFlop to evaluate the balance between cost and performance. On the real-world task of battery material phase segmentation, PaSeR yields a minimum performance improvement of $\mathbf{174}\%$ on the IoU/GigaFlop metric with respect to baselines. We also demonstrate PaSeR's adaptability to complementary models trained on a noisy MNIST dataset, where it achieved a minimum performance improvement on IoU/GigaFlop of $\mathbf{13.4}\%$ over SOTA models. Code and data are available at \url{https://github.com/scailab/paser}.
\end{abstract}

\section{Introduction}
\label{sec:introduction}

\begin{figure}
  \centering
  \includegraphics[width=0.3\textwidth]{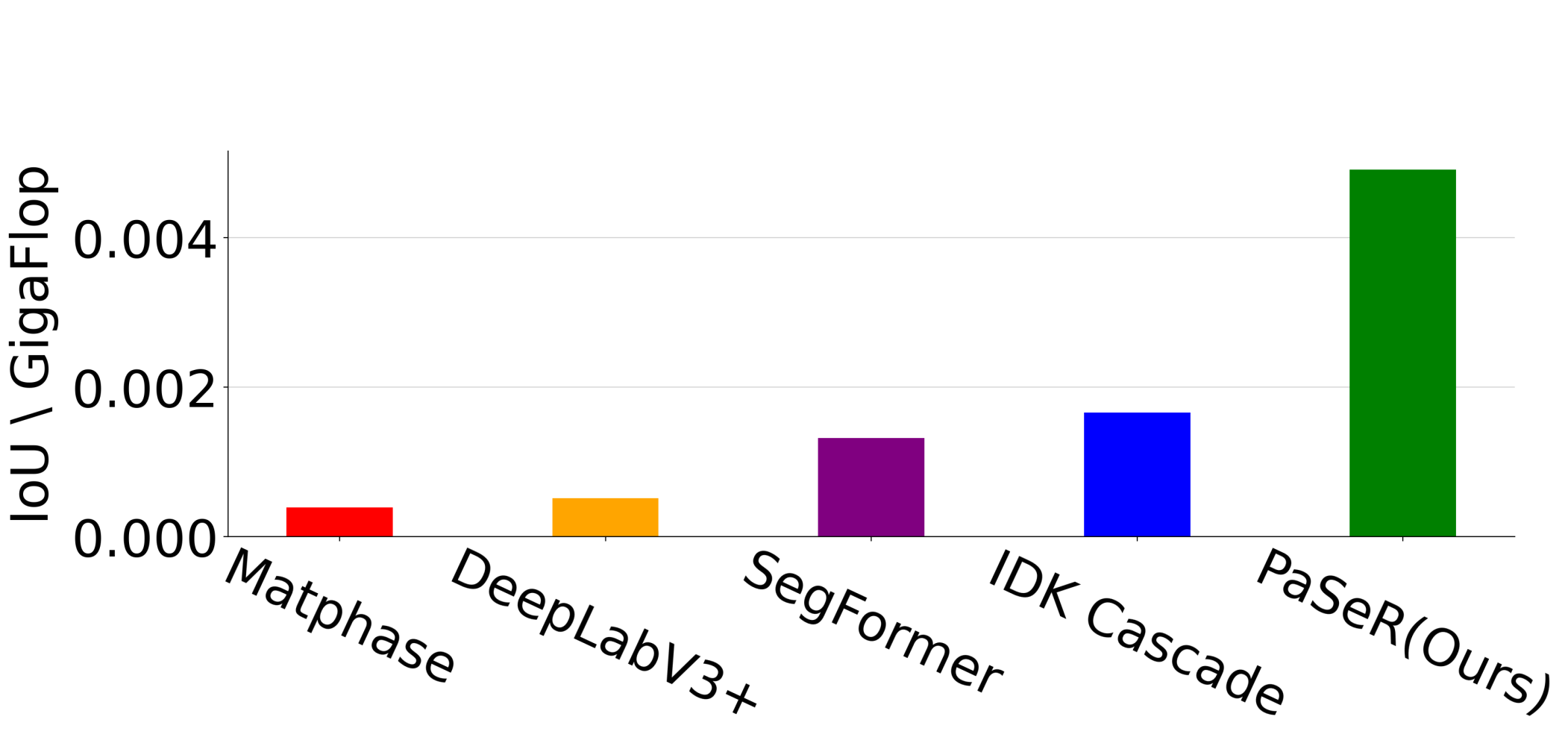}
  \caption{Performance w.r.t IoU/GigaFlop metric (higher is better) of SOTA models and our proposed \gls{ourmethod} model on the battery material phase segmentation task.}
  \label{fig:iou_gf_bar}
\end{figure}

Recent advances in deep learning (DL) and the internet-of-things (IoT) have led to the burgeoning application of DL in manufacturing pipelines~\cite{hussain2020machine,meng2020machine,mohammadi2018deep,tang2017enabling}. In many such applications, ML / DL models are often deployed on devices with low memory and computational capabilities (edge) in conjunction with DL models that are deployed in less constrained environments (fog, cloud). These edge-fog-cloud (EFC) systems are commonly used in areas such as smart manufacturing \cite{chen2018edge} and healthcare \cite{mutlag2021multi} where precision machines such as electrocardiograms collect and preprocess high density data while integrating with a local computer as well as cloud based resources to accurately and efficiently provide critical information.  Although the fog and cloud environments enable the deployment of larger DL models, querying them is costly (due to communication network and model latency). Hence, such real-world contexts require a high-throughput pipeline to balance task accuracy and computational cost.

A popular solution to deal with this problem is the \emph{I Don't Know} (IDK) Cascade~\cite{wang2017idk} in which models of increasing complexity (starting with the least cost model) are sequentially queried until a model yields a prediction exceeding a preset confidence threshold. Multi-exit models \cite{kouris2022multi} follow a similar cascading architecture but require a potentially costly neural architecture search during training. We argue that such pipelines, although well-motivated, lead to high computational costs due to excess computations incurred as a function of the sequential cascading constraint. In this paper, we argue that reinforcement learning (RL) can be employed as an effective substitute to circumvent the \emph{cascading} restriction. We employ RL to directly select which of a set of models to query with a particular input such that the learned policy maximizes task performance while minimizing computational cost. To this end, we propose the \gls{ourmethod} framework and demonstrate its performance on the challenging task of battery material phase segmentation.

\textbf{Application Background}. Lithium-ion batteries are extensively used in many industrial applications, (e.g., smartphones, laptops, and electric vehicles) due to their efficient energy storage capability. The electrode coatings of these batteries consist of composite active materials (e.g., Lithium, Nickel, Manganese) and a polymeric binder (Carbon). The microstructure of these composite electrode coatings consists of the spatial distribution of active and binder materials. The physical parameters of a microstructure, (e.g., homogeneity of coating thickness, porosity) influence battery performance. Resolving the locations of the active and binder materials and their \textit{phase transitions} (i.e., the task of battery material phase segmentation) can help deduce these physical parameters, thereby providing an understanding of phenomena like battery degradation.
Existing techniques to address this problem use expensive high-resolution X-ray computed tomography images~\cite{lu20203d}. Low-resolution (low-res) microtomography images have also been used, but they cannot readily distinguish between spatial distributions of the composite active materials.
Recently, DL segmentation models like \emph{MatPhase} by ~\cite{tabassum2022matphase} have been developed to identify (pixel-wise) these composite materials and their phase transitions from low-res images, however, these approaches are computationally expensive to execute.

In this context, we propose \gls{ourmethod} as a low-cost but effective and robust solution to address the task of battery material segmentation from low-res microtomography images. Our contributions are as follows:
\textbf{(C1)} We develop a novel computationally parsimonious DL framework (employing reinforcement learning with cost-aware rewards) to balance cost with task performance.
\textbf{(C2)} Through qualitative and quantitative experiments, we demonstrate that \gls{ourmethod} yields competitive performance with SOTA models on the battery material phase segmentation task while also being the most computationally efficient.
\textbf{(C3)} We demonstrate the effectiveness of the learned RL policy in unseen (noisy) contexts as well as with task models having complementary strengths.
\textbf{(C4)} Finally, we introduce a novel metric called IoU per GigaFlop (IoU/GigaFlop) which measures the segmentation performance obtained per GigaFlop of computation expended, an effective metric for evaluating such low-cost learning pipelines (see Fig. \ref{fig:iou_gf_bar}).

\section{Related Work}
\label{sec:related_work}

We review two areas of research related to our work, low-cost ML and image segmentation models.

\textbf{Low-Cost \& Tiny ML}. There have been many past efforts to develop low-cost DL pipelines for use in low memory, low storage, high-throughput IoT contexts. \emph{Knowledge distillation} (KD) and employing \emph{decision cascades} are two popular approaches in this context. While the primary goal of KD~\cite{hinton2015distilling,gou2021knowledge,phuong2019towards} is to learn smaller models to \emph{mimic} larger models, this goal isn't fully aligned with the scope of the current work, which is to learn optimal decision pipelines to create a low cost, high performance ML by incorporating multi-models. However, the other research thread of employing decision cascades is directly relevant to our work.  Decision Cascades, originally introduced in~\cite{cai2015learning,angelova2015real} were recently re-popularized by the work of IDK Cascades~\cite{wang2017idk}. The IDK cascade framework imposes a sequential model architecture, where each model is queried in order of increasing complexity until prediction confidence exceeds a threshold. Yet another paradigm of \emph{Tiny-ML}~\cite{Rajapakse_2023,ren2022manage} also aims to develop ML models but with the goal of deploying them on extremely low-cost hardware devices. Our goal is aligned with but complementary to this as our proposed decision pipeline can be employed with such low-cost models along with higher-cost models (on the cloud) to maximize performance and minimize computational cost.

\textbf{Image Segmentation}. The field of image segmentation has also seen many successes in multiple domains ~\cite{chen2017rethinking,li2018pyramid,chen2019learning} with popular architectures like the U-Net~\cite{ronneberger2015unet} and the recent Segment Anything~\cite{kirillov2023segment} \emph{foundation model}. Our \gls{ourmethod} framework is flexible enough to incorporate any of these SOTA segmentation models as we have developed a decision pipeline that can leverage multiple models to maximize performance on a target task while minimizing computational cost. Finally, efforts in intelligent data sampling~\cite{uzkent2020efficient,uzkent2020learning} which may possess a motivation in terms of employing RL for maximal task performance at minimal cost, differ in the actual application of the RL pipeline and learning task.

\section{Problem Formulation}
\label{sec:problem_formulation}
\begin{figure*}[!t]
\centering
    \includegraphics[width=0.65\textwidth]{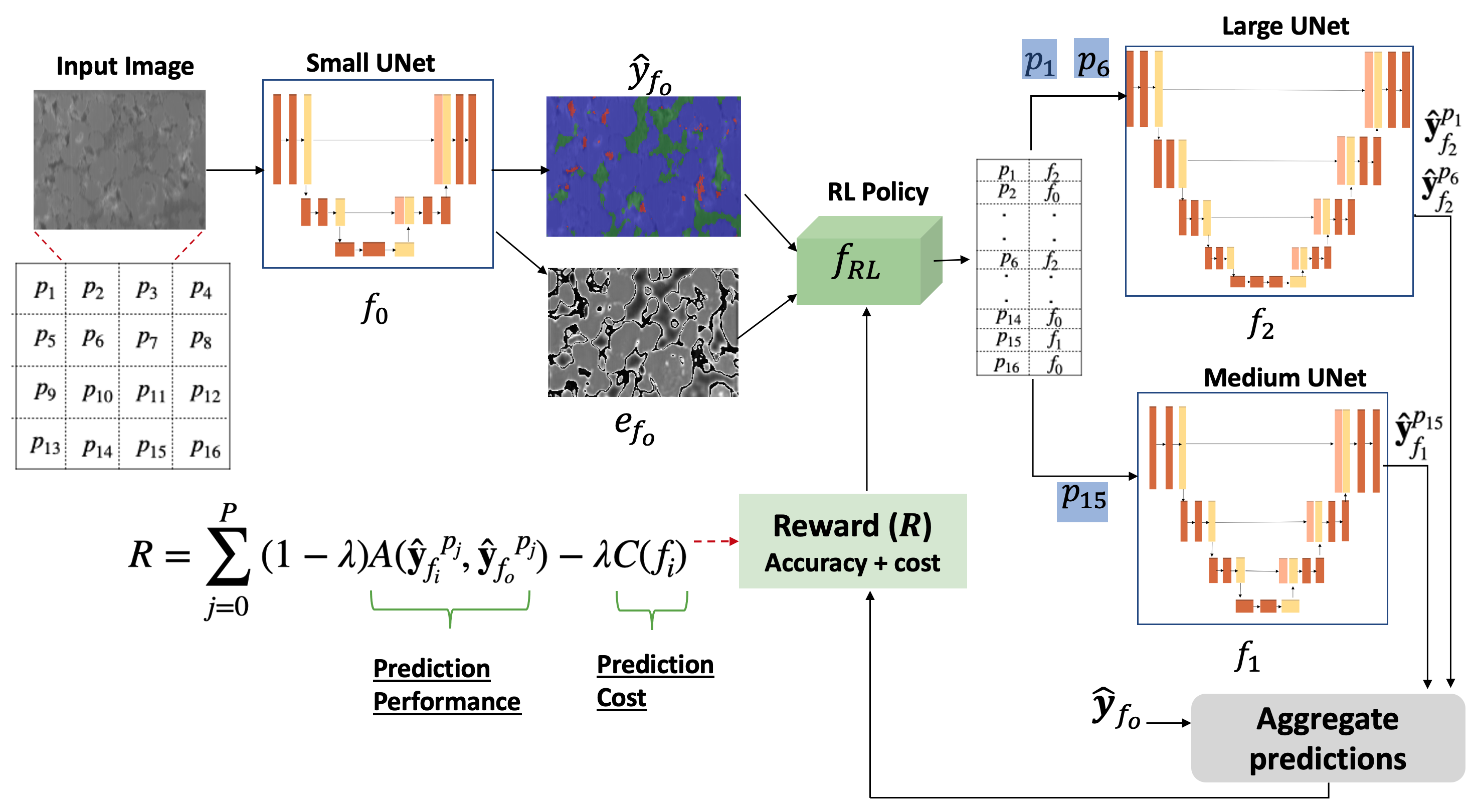}
    \caption{Overview of \gls{ourmethod}. The small UNet ($f_0$) yields the segmentation ($\hat{\mathbf{y}}_{f_0}$) and corresponding entropy map $\mathbf{e}_{f_0}$ conditioned on the whole input image ($\mathbf{x}$). Then, $\mathbf{x}$ is divided into `P' equal sized patches. The RL policy directs each patch $\mathbf{x}^{(p)}$ of $\mathbf{x}$ to one of $f_0, f_1, f_2$ to maximize reward. Based on the RL actions, models $f_1$ and $f_2$ yield predictions for the corresponding image patch. All the predicted patches are then aggregated to yield the final segmentation.}
    \label{fig:paser_overview}
\end{figure*}

In this work, our goal is to develop a learnable decision pipeline that is \emph{computationally parsimonious} (i.e., minimizes wasteful computations) and also yields competitive performance (compared to SOTA models) on the target task. To develop such a decision pipeline, we leverage reinforcement learning (RL). Specifically, we propose the \gls{ourmethod} framework (see Fig.~\ref{fig:paser_overview} for architecture details) composed of an RL policy model $f_{RL}$, a small/efficient task model $f_0$, and $m$ large task models $\{ f_1, \ldots, f_m \}$. In this paper, we demonstrate the performance of \gls{ourmethod} in the context of image segmentation. Algorithm \ref{alg:ours} outlines the training procedure of \gls{ourmethod} in the context of our target task (i.e., image segmentation), but we note that \gls{ourmethod} is task independent and can be applied to other learning contexts with a few appropriate modifications. Code and data are available at \url{https://github.com/scailab/paser}.

\SetKwComment{Comment}{\# }{}
\SetKwInput{KwModels}{Models}
\SetKwInput{KwHyp}{Hyp}
\SetKwInput{KwRepeatLines}{Repeat Lines}
\RestyleAlgo{ruled}
\begin{algorithm*}[ht]
    \footnotesize
    \caption{\gls{ourmethod} Algorithm}\label{alg:ours}
    \KwData{$\mathcal{D}_{PT}, \mathcal{D}_{RL}, \mathcal{D}_{FT} \quad \textbf{Parameters: } \btheta_{f_{RL}}, \btheta_{f_{0}}, \ldots, \btheta_{f_m} \quad \textbf{Hyp: } \lambda, \eta, \beta$}
    \KwModels{RL policy $f_{RL}$, small/efficient model $f_0$ and $m$ large task models $\{ f_1, \ldots, f_m \}$}
    \For{$f_i \in \{f_1, \ldots, f_m\}$ \hspace*{17ex}\Comment{Pretrain each large task model}}{
        \For{$\bx^{(p)}, \by^{(p)} \in \mathscr{P}(\mathcal{D}_{PT})$ \hspace*{9.6ex}\Comment{For each data point in pre-training dataset}}{
            $\hat{\by}_{f_i}^{(p)}, \hat{\bz}_{f_i}^{(p)} \gets f_i(\bx^{(p)})$ \hspace*{9.6ex}\Comment{Get task predictions and logits from model $f_i$}
            $l \gets \mathcal{L}(\hat{\bz}_{f_i}^{(p)}, \by^{(p)})$ \hspace*{12.73ex}\Comment{Compute loss (cross entropy)}
            $\btheta_{f_i} \gets \btheta_{f_i} - \eta \nabla_{\btheta_{f_i}} l$ \hspace*{11.2ex}\Comment{Update model parameters}
        }
    }
    \For{$\bx, \by \in \mathcal{D}_{PT}$ \hspace*{23ex}\Comment{Pretrain small/efficient model with KD loss}}{
        $\hat{\by}_{f_0}, \hat{\bz}_{f_0} \gets f_0(\bx)$ \hspace*{17ex}\Comment{Get \emph{small} model ($f_0$) prediction}
        $\hat{\by}_{f_m}, \hat{\bz}_{f_m} \gets f_m(\bx)$ \hspace*{15.0ex}\Comment{Get largest model prediction}
        $l \gets \mathcal{L}(\hat{\bz}_{f_0}, \by) + \beta \mathcal{L}_{KD}(\hat{\bz}_{f_0}, \hat{\bz}_{f_m})$ \hspace*{1.4ex}\Comment{Compute loss with KD}
        $\btheta_{f_0} \gets \btheta_{f_0} - \eta \nabla_{\btheta_{f_0}}l$ \hspace*{15.1ex}\Comment{Update model parameters}
    }
    \For{$\bx, \by \in \mathcal{D}_{RL}$ \hspace*{23.6ex}\Comment{Train RL policy model}}{
        $\hat{\by}_{f_0}, \be_{f_0} \gets f_0(\bx)$ \hspace*{17.2ex}\Comment{Get small model prediction and entropy}
        $\bs \gets f_{RL}(\hat{\by}_{f_0}, \be_{f_0})$ \hspace*{15.8ex}\Comment{Get probabilities of actions from RL model}
        $\ba \sim \pi_{RL}(\mathcal{A} \mid \bs)$ \hspace*{18.62ex}\Comment{Sample action from RL policy distribution}
        \For{$a_p \in \ba$ \hspace*{24.3ex}\Comment{For each model and patch in action}}{
            $\hat{\by}_{a_p}^{(p)} \gets f_{a_p}(\bx^{(p)})$ \hspace*{24.0ex}\Comment{Get model prediction}
            $R \pluseq (1-\lambda)A(\hat{\by}_{f_{a_p}}^{(p)}, \hat{\by}_{f_0}^{(p)}) - \lambda C(f_{a_p})$ \hspace*{3ex}\Comment{Compute accuracy+cost-based reward}
        }
        $\nabla_{\btheta_{RL}} J = \mathbb{E}[\nabla_{\btheta_{RL}} \log \pi_{RL}(\mathcal{A}|s) * R]$ \hspace*{9.3ex}\Comment{Compute policy gradient}
        $\btheta_{f_{RL}} \gets \btheta_{f_{RL}} - \eta \nabla_{\btheta_{f_{RL}}} J(\pi_{RL})$ \hspace*{14ex}\Comment{Update RL model}
    }
    \For{$\bx, \by \in \mathcal{D}_{FT}$ \hspace*{23.5ex}\Comment{Finetune models}}{
        \KwRepeatLines{5-7 for each large model}
        \KwRepeatLines{19-28 for RL model}
    }
\end{algorithm*}

\textbf{Segmentation Model Pretraining}. At the outset of our training procedure, we split the training data into three equal subsets: $\mathcal{D}_{PT}, \mathcal{D}_{RL}, \mathcal{D}_{FT}$. Each subset is comprised of image instances and pixel labels $(\bx, \by)$ where $\bx \in \mathbb{R}^{C \times H \times W}$ and $\by \in \mathbb{R}^{1 \times H \times W}$. Using the pretraining subset $\mathcal{D}_{PT}$, we train the $m$ large segmentation models $f_1, \ldots, f_m$ first by splitting each image $\mathbf{x}$ into $P$ equal size patches (in our case $P = 16$) with the help of a \emph{patchification} function $\mathscr{P}(\cdot)$ where $\mathbf{x}^{(p)}$ denotes the $p^{th}$ patch. These patches are passed as inputs to each model while optimizing cross entropy loss $\mathcal{L}(\hat{\bz}^{(p)}, \by^{(p)})$ between the prediction logits $\hat{\bz}^{(p)}$ and ground truth $\by^{(p)}$.
Once models $f_1, \dots, f_m$, are pre-trained, the smallest model, $f_0$ is pre-trained using $\mathcal{D}_{PT}$ on the full image (i.e., no patchification). In addition to using the cross entropy loss $\mathcal{L}(\hat{\bz}, \by)$ we also use a knowledge distillation (KD) loss \citep{hinton2015distilling,kim2021comparing} between the outputs of the largest model $f_m$ and the small model $f_0$. We define the KD loss function in Eq.~\ref{eq:kd_loss}. 
\begin{equation}
    \label{eq:kd_loss}
    \mathcal{L}_{KD} = \frac{1}{|\mathcal{D}_{PT}|} \sum_{j=1}^{|\mathcal{D}_{PT}|} \left(\hat{\bz}^{(p)}_{{f_0}, j} - \hat{\bz}^{(p)}_{{f_m}, j}\right)^2
\end{equation}
The term $\hat{\mathbf{y}}^{(p)}_{f_0,j}$ indicates the segmentation predictions for patch $p$ of instance $j$ yielded by model $f_0$. $\hat{\mathbf{y}}^{(p)}_{f_m,j}$ is the corresponding prediction yielded by model $f_m$.
This loss encourages outputs of $f_0$ to be closer to the largest model $f_m$, thereby transferring information from the representations learned by $f_m$ to $f_0$ improving its performance without increasing its size.

\textbf{RL Training}. We incorporate reinforcement learning as the decision paradigm to develop a compute-efficient segmentation pipeline. Specifically, our RL policy is conditioned upon states $\mathbf{s}$, constituted by the image segmentation $\hat{\mathbf{y}}_{f_0}$ and entropy maps $\be_{f_0}$ of the smallest model $f_0$ to output an action which specifies a set of patch and model pairs for each image to be passed \emph{upstream} to more sophisticated models in the pipeline. States are of the form $(\hat{\by}_{f_0}, \be_{f_0})$ and actions are defined as $\ba \in \{0, \ldots, m\}^P$. We define the patch-model selection policy as ${\pi_{RL}(\ba \mid \bs) = p(\ba \mid f_{RL}(\hat{\by}_{f_0}, \be_{f_0}; \btheta_{f_{RL}}))}$.
Here the policy network $f_{RL}$ parameterizes the action distribution $p$, which in our case is a categorical distribution with probabilities ${\mathbf{s} \in \{ s_{f_0}, \ldots, s_{f_m} \}^{P} : s_{f_i} > 0,\,\, \sum_{i = 0}^{m} s_{f_i} = 1}$.
The entropy $\be_{f_0}$ is calculated using Monte Carlo dropout \cite{gal2016dropout}, but note that other methods for uncertainty quantification can also be supported by \gls{ourmethod}.

Using probabilities $\mathbf{s}$, we sample from a categorical distribution to obtain an action $\mathbf{a} \in \{ 0, \ldots, m \}^{P}$. For example, if $\mathbf{a}_k$ = 2 for some index $k$ of $\mathbf{a}$, this indicates that $f_{RL}$ has chosen  the $k^{th}$ patch to be directed to model $f_2$ for segmentation. 
Using the sampled action, we pass each patch to its respective model and compute a reward. The reward function is detailed in Eq.~\ref{eq:reward} and is based on the difference in prediction performance $A$ between the large and small model predictions, $\hat{\mathbf{y}}_{f_{a_p}}, \hat{\mathbf{y}}_{f_0}$, as well as a computational cost penalty term $C$. The action $\ba$ defines the models run on each patch.
\small
\begin{equation}
    R(\ba = \{a_1, \ldots, a_P\}) = \sum_{p=0}^P (1 - \lambda) A(\hat{\by}_{f_{a_p}}^{(p)}, \hat{\by}_{f_0}^{(p)}) - \lambda C(f_{a_p})
    \label{eq:reward}
\end{equation}
\normalsize
For our experiments in segmentation we use the difference in mean intersection over union (IoU) as our measure of prediction performance: ${A(\hat{\by}_{f_i}^{(p)}, \hat{\by}_{f_0}^{(p)}) = IoU(\hat{\by}_{f_i}^{(p)}) - IoU(\hat{\by}_{f_0}^{(p)})}$.
Note that in Eq.~\ref{eq:reward}, the cost parameter $\lambda$ parameterizes a convex combination of accuracy and computational cost to provide a simple way to control the influence of each component on the RL policy reward.
We design a cost function $C$ in Eq. \ref{eq:cost_func} with range $(0, 1)$ as the ratio of the number of learnable parameters in a model to the total number of parameters in all models $\{f_0,\dots,f_m\}$.
\small
\begin{equation}
    C(f_i) = \frac{\text{numParams}(f_i)}{\sum_{j=1}^m \text{numParams}(f_j)}
    \label{eq:cost_func}
\end{equation}
\normalsize
Using the reward value $R$, we compute the policy gradient \citep{sutton1999policy} $\nabla_{\btheta_{f_{RL}}} J = \mathbb{E}[\nabla_{\btheta_{f_{RL}}} \log \pi_{RL}(\ba \mid \bs) * R]$ and update the parameters $\btheta_{RL}$ of the RL policy.

\textbf{Fine-Tuning}. The final step of \gls{ourmethod} is fine-tuning. Here, we jointly update the large models and RL model. The joint training helps the large segmentation models improve their performance on the inputs being directed to them by the RL policy while also further personalizing the RL policy to discern the strengths and weaknesses of each constituent segmentation model for each input patch.

\section{Experimental Setup}
\label{sec:experimental_setup}
We train three UNet segmentation models $f_0, f_1, f_2$ with 16571, 1080595, and 17275459 parameters respectively on $\mathcal{D}_{PT}$ for 200 epochs, followed by training our RL model $f_{RL}$ with 14736 parameters on $\mathcal{D}_{RL}$ for 200 epochs. Finally, we fine-tune all models on $\mathcal{D}_{FT}$ for 200 epochs. \gls{ourmethod} trains with a batch size of 32 using the Adam optimizer \citep{kingma2014adam} with $\eta = 0.0001$. In our battery segmentation experiment we set $\beta = 0.01$ using grid search and $\lambda = 0.5$ which corresponds to an even balance between performance and cost. For the noisy MNIST dataset we set $\lambda = 0$, see section R4. Adaptability to Complementary Models for more details.

\subsection{Baselines}\label{sec:baselines}
We compare \gls{ourmethod} to six baselines with complementary strengths to illustrate how we improve upon each of these baselines in either IoU performance and/or IoU per GigaFlop efficiency.
\textbf{(1) \gls{idkbaseline}~\cite{wang2017idk}}:
We implement the \gls{idkbaseline} model with a cost aware cascade using the same segmentation models in \gls{ourmethod}. For the IDK loss and cost function we use cross entropy loss and our previously defined cost function (Eq. \ref{eq:cost_func}), while tuning this baseline with an exhaustive grid search.
\textbf{(2) \gls{ourmethodrandpolicy}}:
We setup \gls{ourmethod} with a random policy for actions drawn uniformly from a categorical distribution. We call this method \gls{ourmethodrandpolicy}
\textbf{(3) MatPhase~\cite{tabassum2022matphase}}:
We also compare \gls{ourmethod} to a state of the art (SOTA) model specialized for the task of battery material phase segmentation. The MatPhase model is an ensemble method which combines UNet segmentation models with pixel level IDK classification and a convolutional neural network.
\textbf{(4) DeepLabV3+ \cite{chen2018encoder}}:
To put \gls{ourmethod} in context with modern DL models, we compare it to DeepLabV3+, a SOTA segmentation model which uses atrous convolutions alongside an encoder-decoder.
\textbf{(5) SegFormer \cite{xie2021segformer}}:
We also compare our method to SegFormer, a recent SOTA segmentation model which combines transformers with small multi-layer perceptron decoders.
\textbf{(6) EfficientViT \cite{cai2022efficientvit}}: We also compare to the lightweight EfficientViT, which uses linear attention.

\subsection{Evaluation Metrics}\label{sec:eval_metrics}
\textbf{(1) Intersection-Over-Union$\,$(IoU)}: We employ  IoU (aka. Jaccard index), a popular and effective metric used to evaluate performance on image segmentation tasks.
\textbf{(2) Flops (F)}: We profile the number of floating point operations per instance for \gls{ourmethod} and baselines in inference mode when run on the full test set. This gives us the raw computational cost of each model.
\textbf{(3) IoU Per GigaFlop \big($\,\frac{\mathrm{IoU}}{\mathrm{GigaFlop}}\,$\big)}: While Flops measures compute required per model, we introduce a new metric called IoU per GigaFlop which is defined by the ratio $\frac{\text{IoU}}{\text{GigaFlop}}$. This metric enables a unified understanding of performance effectiveness and computational cost.

\subsection{Dataset Description}
\par \noindent
\textbf{Battery Material Phase Segmentation}. Our battery material phase segmentation dataset consists of 1,330 images (1270 training images, 20 validation, and 40 test images) obtained from low-res microtomography (inputs), each of size $(224, 256)$ along with pixel level labels of 3 classes (obtained from high-res computational tomography): pore, carbon, and nickel. We split these images into 16 equal size patches of size $(56, 64)$ each.

\par \noindent
\textbf{Noisy MNIST}. The standard MNIST dataset \cite{deng2012mnist} consists of 70,000 grayscale images (50,000 training, 10,000 validation and 10,000 test images). We create three different versions of this dataset for foreground/background segmentation with three noise types respectively: Gaussian blur with radius 1, Gaussian blur with radius 2 and a box blur with a fixed convolutional filter. See Fig.~\ref{fig:mnist_noise_types} for examples of each noise type.

\section{Results \& Discussion}
\label{sec:results}

\begin{table*}
  \centering
  \sisetup{table-alignment-mode = format,
table-number-alignment = center}
  \small
  \begin{tabular}{@{}
  l
  c
  S[table-format = 2.2e2]
  S[table-format = 2.2e2]|
  c
  S[table-format = 2.2e2]
  S[table-format = 2.2e2]
  @{}}
    \toprule
    \multirow{2}{*}{Model} &
      \multicolumn{3}{c}{\underline{Battery}} &
      \multicolumn{3}{c}{\underline{Noisy MNIST}} \\
      & {IoU} & {Flops} & {IoU/GigaFlop} & {IoU} & {Flops} & {IoU/GigaFlop} \\
      \midrule
    Matphase~\cite{tabassum2022matphase} & \num{0.8144} & 2.11e+12 & 0.39e-3 & {\longdash[2]} & {\longdash[2]} & {\longdash[2]} \\
    DeepLabV3+~\cite{chen2018encoder} & \num{0.7817} & 1.55e+12 & 0.51e-3 & \num{0.8459} & 2.07e+13 & 4.08e-05 \\
    SegFormer~\cite{xie2021segformer} & \num{0.7692} & 5.84e+11 & 1.32e-3 & \num{0.8448} & 7.56e+12 & 1.12e-4 \\
    EfficientViT~\cite{cai2022efficientvit} & \num{0.7765} & 4.34e+11 & 1.79e-3 & \num{0.8344} & 3.72e+14 & 2.24e-06 \\
    \gls{idkbaseline}~\cite{wang2017idk} & \num{0.6987} & 4.20e+11 & 1.66e-3 & \num{0.7750} & 1.15e+13 & 6.73e-05 \\
    \gls{ourmethodrandpolicy} & \num{0.7234} & 5.33e+11 & 1.36e-3 & \num{0.6376} & 7.05e+12 & 9.05e-05 \\
    \gls{ourmethod} (ours) & \num{0.7426} & 1.51e+11 & \bfseries 4.91e-3 & \num{0.8231} & 6.51e+12 & \bfseries 1.27e-4 \\
    \bottomrule
  \end{tabular}
  \caption{Battery material phase segmentation and Noisy MNIST results comparison between \gls{ourmethod} and SOTA models.}
  \label{tab:main_results}
\end{table*}

\begin{table*}
  \centering
  \sisetup{table-alignment-mode = format,
table-number-alignment = center}
  \small
  \begin{tabular}{@{}
  l
  c
  S[table-format = 2.2e2]
  S[table-format = 2.2e2]|
  c
  S[table-format = 2.2e2]
  S[table-format = 2.2e2]
  @{}}
    \toprule
    \multirow{2}{*}{Model} &
      \multicolumn{3}{c}{\underline{Battery}} &
      \multicolumn{3}{c}{\underline{Noisy MNIST}} \\
      & {IoU} & {Flops} & {IoU/GigaFlop} & {IoU} & {Flops} & {IoU/GigaFlop} \\
      \midrule
    \gls{idkbaselineioumatch} & \num{0.7444} & 1.54e+12 & 0.48e-3 & \num{0.7755} & 1.15e+13 & 6.74e-05 \\
    \gls{ourmethod} (ours) & \num{0.7426} & 1.51e+11 & \bfseries 4.91e-3 & \num{0.8231} & 6.51e+12 & \bfseries 1.27e-4 \\
    \bottomrule
  \end{tabular}
  \caption{Battery material phase segmentation and Noisy MNIST IoU Match Results for \gls{ourmethod} and \gls{idkbaseline}.}
  \label{tab:iou_match_results}
\end{table*}

In line with our goal of designing a computationally parsimonious framework, we investigate \gls{ourmethod} performance in the context of the following research questions. 

\par \noindent 
\textbf{R1.} How does the task performance and computational efficiency of \gls{ourmethod} compare with the \gls{idkbaseline} paradigm?

\par \noindent 
\textbf{R2.} How well does \gls{ourmethod} balance IoU and efficiency relative to SOTA segmentation models?

\par \noindent
\textbf{R3.} How adaptable and robust is the \gls{ourmethod} decision policy to noisy data? 

\par \noindent
\textbf{R4.} How adaptable and robust is the \gls{ourmethod} decision policy to task models with complementary strengths?

\par \noindent
\textbf{R5.} What are the effects of the various components of \gls{ourmethod}, ($\lambda$, MC-Sampling) on achieving an effective balance between computational cost and task performance?

\subsection{R1: Task Performance and Computational Efficiency vs. \texorpdfstring{\gls{idkbaseline}}{IDK-Cascade}}

To evaluate model task performance, we compare our \gls{ourmethod} model to the cost-aware IDK cascading decision baseline, and a variant of \gls{ourmethod} (i.e., \gls{ourmethodrandpolicy}) with the same segmentation models as \gls{ourmethod} except with a random policy instead of a learned RL policy. The performance results are depicted in Table~\ref{tab:main_results}. Looking at the battery dataset, we see that \gls{ourmethod} outperforms the IDK Cascade model by $\mathbf{6.28}\%$ in terms of the IoU metric. \gls{ourmethod} also achieves the highest IoU/GigaFlop, outperforming \gls{idkbaseline} by $\mathbf{196}\%$.

Note that the \gls{idkbaseline} model currently under-performs~\gls{ourmethod} on the Battery dataset. Hence, for a fair comparison with our method, we tune the IDK Cascade model to \emph{match} the IoU performance of \gls{ourmethod} and denote this model as \gls{idkbaselineioumatch}. We achieve this by adjusting the entropy thresholds used in each stage of the cascade until we obtain a least-upper-bound performance (i.e., within a tolerance of $10^{-3}$ of IoU) compared to \gls{ourmethod} on the same test set. In Table \ref{tab:iou_match_results}, comparing the flops of both models (for the same IoU performance), we see that the \gls{ourmethod} model requires \textbf{90}\% fewer flops compared to \gls{idkbaselineioumatch} to achieve similar performance. This is further corroborated by the IoU/GigaFlop metric in Table~\ref{tab:iou_match_results} wherein we see that \gls{ourmethod} achieves a \textbf{923}\% improvement on this metric thereby indicating that \gls{ourmethod} is able to yield good performance at much lower computational cost compared to the IDK cascading modeling paradigm.

Finally, on the MNIST dataset \gls{ourmethod}  outperforms \gls{idkbaselineioumatch} by $\mathbf{6.1}\%$ and $\mathbf{88.4}\%$ on IoU and IoU/GigaFlop metrics respectively. Here \gls{idkbaselineioumatch} underperforms on the IoU metric vs \gls{ourmethod} because the entropy based threshold of \gls{idkbaselineioumatch} is not nuanced enough to determine the correct model assignment for a given input. In fact, the accuracy of model assignment by the \gls{idkbaselineioumatch} is only $80\%$ while \gls{ourmethod} has a model assignment accuracy of $92.7\%$.

\subsection{R2: Performance Comparison with SOTA Segmentation Models}

The problem of battery material phase segmentation has been investigated by a few previous efforts (see Sec.~\nameref{sec:related_work}). The most recent and best model of this group of efforts is MatPhase. We characterize the performance of \gls{ourmethod} with respect to this SOTA battery material phase segmentation model as well as the recent monolithic SOTA segmentation models DeepLabV3+, SegFormer and EfficientViT. The distributed nature of \gls{ourmethod} vs monolithic architectures such as SegFormer allows PaSeR to be deployed in an EFC system where monolithic SOTA models would not satisfy computational edge constraints.

In Table \ref{tab:main_results} we see that although MatPhase ~\cite{tabassum2022matphase} outperforms \gls{ourmethod} in terms of segmentation performance, it does so employing significantly more computation. Specifically, MatPhase employs \textbf{1297}\% more computation than \gls{ourmethod} to obtain a \textbf{9.7}\% performance improvement.
Further, we notice that \gls{ourmethod} achieves a minimum improvement of \textbf{174}\% over all baselines on the IoU/GigaFlop metric.
This is a significant result showing the usefulness of \gls{ourmethod} relative to SOTA models like MatPhase in computationally constrained environments.

When comparing to DeepLabV3+, SegFormer and EfficientViT on the Battery dataset, we see that \gls{ourmethod} is within $4\%$ of the IoU that those models achieve. Despite their slightly better performance on IoU, \gls{ourmethod} is much more efficient on the IoU/GigaFlop metric by $\mathbf{863}\%$, $\mathbf{272}\%$ and $\mathbf{174}\%$ for DeepLabV3+, SegFormer and EfficientViT respectively. On the Noisy MNIST dataset, we see the same pattern again. For the DeepLabV3+ model, \gls{ourmethod} has an $\mathbf{211}\%$ higher IoU/GigaFlop while also outperforming the SegFormer model by $\mathbf{13.4}\%$ on IoU/GigaFlop. The EfficientViT model performs poorly on this dataset because it is designed for high resolution images and downscales the image by a factor of 8 when outputting segmentation maps. To compensate for this downscaling, we upscale our 32x32 MNIST images to 256x256 for this model.

\textbf{Cityscapes.}
To demonstrate \gls{ourmethod} on a modern segmentation task while also integrating pretrained models, we train \gls{ourmethod} on the Cityscapes dataset \cite{cordts2016cityscapes} using three task models: our small UNet, SegFormer-B0, and SegFormer-B5 with $\lambda = 0.10$ achieving a test set IoU of 0.8163 which is comparable with SOTA model performance.

\subsection{R3: Adaptability to Unseen Contexts (Battery Data)}

\begin{figure}[t]
  \centering
  \includegraphics[width=0.34\textwidth]{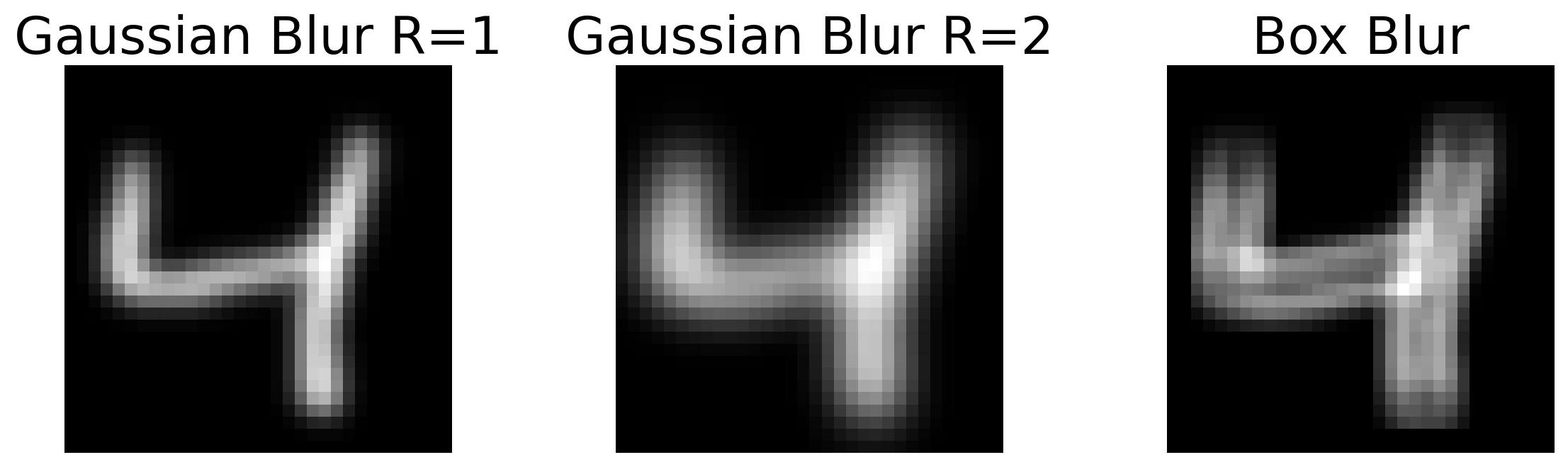}
  \caption{Examples of types of noise added to MNIST data.}
  \label{fig:mnist_noise_types}
\end{figure}

\begin{figure}
  \centering
  \includegraphics[width=0.47\textwidth]{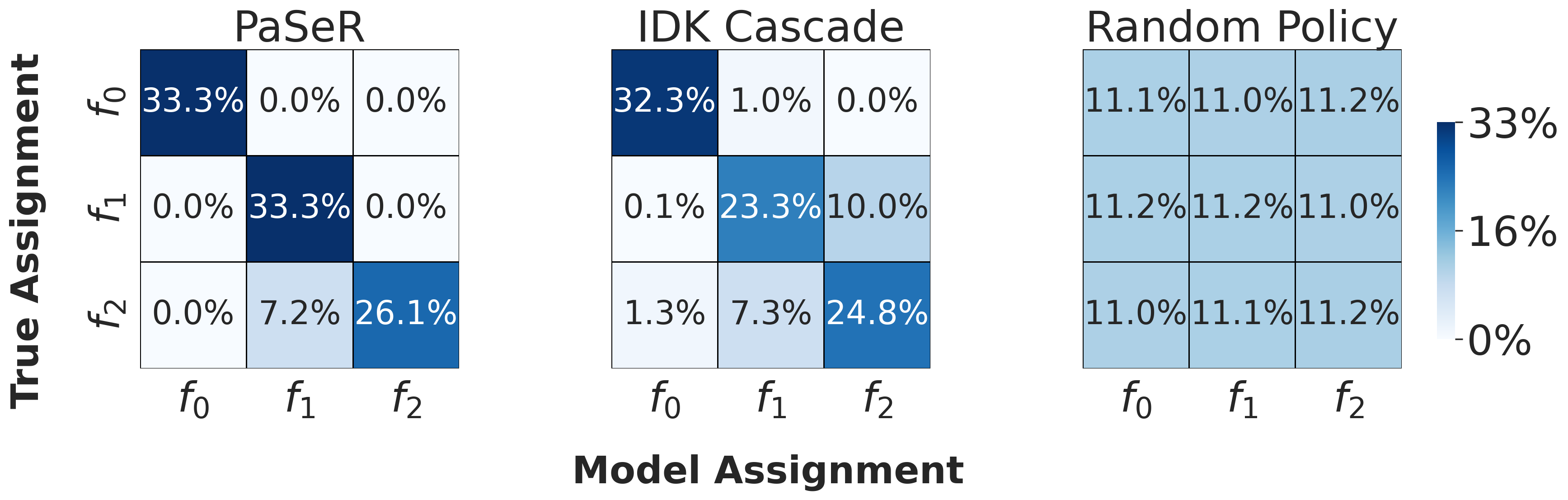}
  \caption{Model assignment confusion matrices for \gls{ourmethod}, \gls{idkbaseline} and \gls{ourmethodrandpolicy}}
  \label{fig:conf_mats}
\end{figure}

Data and products in real-world (IoT-based) manufacturing pipelines are often plagued by process noise leading to instances from unseen input data distributions. It is in such contexts that the true effectiveness of pipelines such as \gls{ourmethod} come to the fore in terms of being able to adapt in unseen data contexts.

\begin{table}[t]
  \centering
   \small
  \begin{tabular}{lccc}
    \toprule
    Model & IoU (Noisy) & Degradation \\
    \midrule
    \gls{ourmethodrandpolicy} & 0.5864 & -18.94\% \\
    \gls{ourmethod} & 0.7322 & -1.4\% \\
    \bottomrule
  \end{tabular}
  \caption{\gls{ourmethod} vs~\gls{ourmethodrandpolicy} on noisy datasets. Note that \gls{ourmethodrandpolicy} fails to adapt in the case of noisy data.}
  \label{tab:rand_vs_ours}
\end{table}

To investigate the adaptability of our RL policy based \gls{ourmethod} and demonstrate its effectiveness relative to the random policy in \gls{ourmethodrandpolicy}, 
we create a variant of our battery segmentation dataset injected with salt and pepper noise. This is done to simulate data quality degradation of the input to the segmentation pipeline, due to equipment / process noise. Further, we create pre-trained variants of all segmentation models  $\{f_1,\dots,f_m\}$ (except $f_0$ i.e., the small U-Net) on a combination of clean and noisy data. Finally, we just replace (without fine-tuning $f_0$, $f_{\mathrm{RL}}$) the models $\{f_1,\dots,f_m\}$ in the fully-trained \gls{ourmethod} model, with variants trained on noisy as well as clean data.

We then investigate performance of \gls{ourmethod} and \gls{ourmethodrandpolicy} (both augmented with same set of segmentation models) on a noisy held-out set of data. Note that by leaving $f_0$ and RL policy $f_{\mathrm{RL}}$ unaware of the noisy data, we have created a scenario which is unseen w.r.t the RL policy (and model $f_0$ on whose predictions and entropy the RL policy decisions are conditioned). 

Table~\ref{tab:main_results} showcases IoU segmentation results (on the battery dataset) of \gls{ourmethod} and \gls{ourmethodrandpolicy} in the clean data context while Table~\ref{tab:rand_vs_ours} showcases corresponding IoU results in a noisy context. From these results, we notice that both models experience degradation under the unseen noisy context. However, the degradation in IoU performance experienced by \gls{ourmethod} is minimal (1.4\%), owing to the RL policy being able to adapt, unlike in \gls{ourmethodrandpolicy} which shows significant performance degradation (18.94\%).
We find that \gls{ourmethod} sends 5.7\% more patches to the larger models (that have been exposed to the noisy data) than in the clean data case, thereby showcasing strong evidence of adaptability in unseen contexts. This \textbf{advantage of adaptability in noisy, unseen scenarios with minimal degradation} is also a significant advantage of \gls{ourmethod} and its cost-aware RL model.

\subsection{R4. Adaptability to Complementary Models (Noisy MNIST)}
\label{sec:expt_complement}
We demonstrate robustness of \gls{ourmethod} to utilize models with complementary strengths, on the Noisy MNIST dataset. We train each segmentation model $(f_0, f_1, f_2)$ on the task of foreground/background segmentation on each noisy dataset respectively, training $f_0$ on the Gaussian blur with radius 1, $f_1$ on Gaussian blur radius 2 and $f_3$ on box blur data. Examples of the three noise types are shown in Fig. \ref{fig:mnist_noise_types}. Each segmentation model learns how to denoise its own noise type and thereby has a unique strength relative to other models.

After training the segmentation models, we train \gls{ourmethod}'s RL policy with $\lambda = 0$ such that it learns the optimal policy without regard for computational cost. We have the dataset containing equal proportions of each noise type, so the optimal policy will send one-third of the images to each segmentation model. Then we fine-tune the pre-trained RL model assuming it has learned an optimal policy. We do this by linearly increasing $\lambda$ while measuring the total variation distance (TVD) from the optimal policy which was previously learned. Once this TVD hits a pre-specified threshold, we stop fine-tuning.

To understand the robustness of the \gls{ourmethod} RL policy, we examine the model assignment confusion matrices in Fig. \ref{fig:conf_mats}. Here, \gls{ourmethod} (with a TVD threshold of 10\%) has nearly perfect assignment of images to the $f_0$ and $f_1$ task models, while only sending 7.2\% of images which should have gone to the $f_2$ model to the $f_1$ model. This occurs because of the 10\% TVD threshold, which gives \gls{ourmethod} the flexibility to send a small percentage of images to the $f_1$ model instead of $f_2$. Comparing this to the model assignment of \gls{idkbaseline}, we see that it sends 10\% of $f_1$ model images to $f_2$, while also incorrectly sending 7.2\%  of $f_2$ model images to $f_1$. This is why \gls{idkbaseline} cannot match the performance of \gls{ourmethod}. The \gls{idkbaseline} with entropy as the gating mechanism is not adaptable enough to accurately assign images to the best model. Finally, note that the \gls{ourmethodrandpolicy} assigns images at random to each task model and thereby has the poorest performance across all metrics.

\subsection{R5: Sensitivity to Hyperparameters}

\begin{figure}
  \centering
  \includegraphics[width=0.47\textwidth]{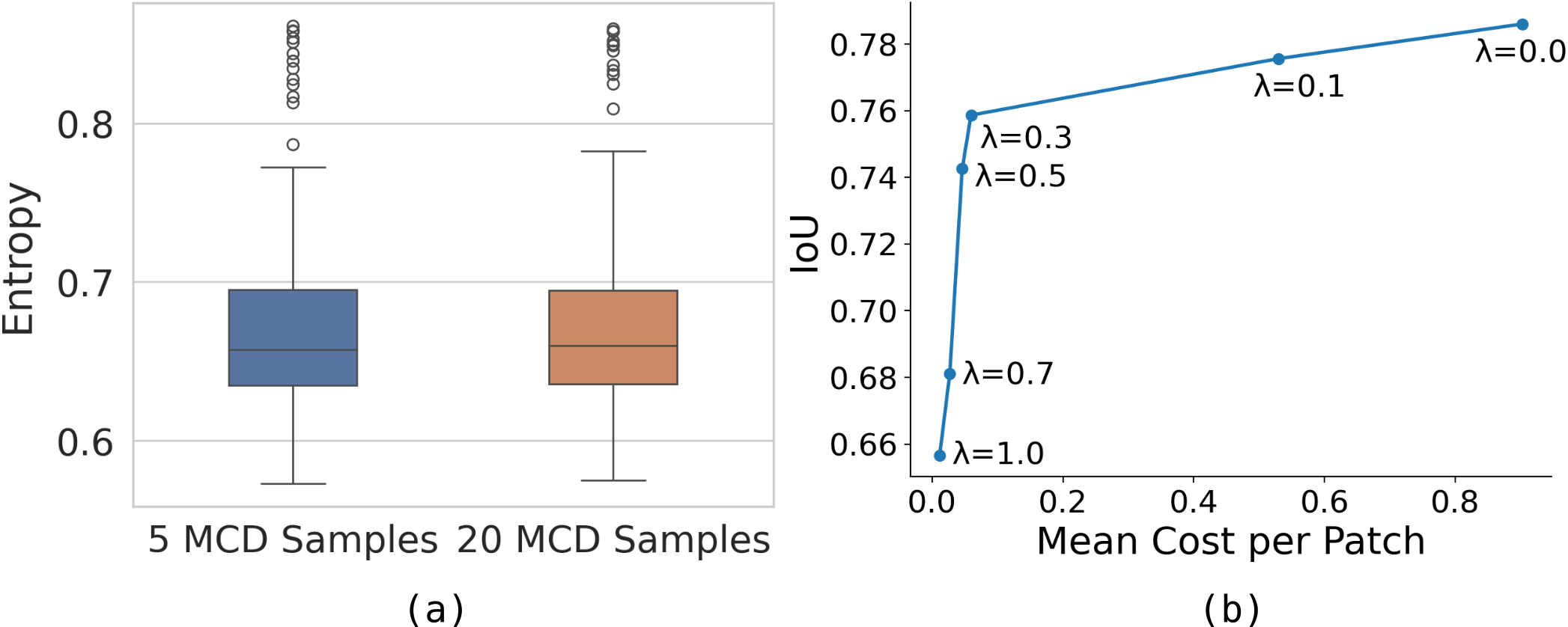}
  \caption{(a) Distribution of entropy estimates with 5 and 20 Monte Carlo Dropout (MCD) samples. (b) \gls{ourmethod} IoU vs Mean Cost as $\lambda$ changes on battery material phase segmentation dataset.}
  \label{fig:mcd_iou_cost}
\end{figure}

We now investigate how $\lambda$ (cost parameter) and entropy map estimation affect \gls{ourmethod} performance.
\par \noindent
\textbf{Performance vs Cost Trade-off.}
In Fig. \ref{fig:mcd_iou_cost}(b), we show \gls{ourmethod}'s performance/cost trade-off curve as $\lambda$ decreases for the battery segmentation task. The mean cost is calculated using Eq.~\ref{eq:cost_func}. This cost function is based on the number of parameters in each task model with $f_2$ having a significantly higher cost than $f_1$. As expected, as $\lambda$ increases, mean cost falls and performance decreases. The sharp drop in cost between $\lambda = 0.0$. and $\lambda = 0.3$ occurs because of the high difference in the cost of using the large task model $f_2$ vs using the smaller models. As $\lambda$ increases in this range, \gls{ourmethod} uses $f_2$ less, leading to a quick drop in mean cost.
\par \noindent
\textbf{Effect of Number of MCDropout Samples.}
\gls{ourmethod} computes entropy maps using Monte Carlo (MC) dropout sampling which requires taking multiple samples of each prediction. To test the sensitivity of estimation of entropy to the number of MC samples taken, we show a box plot of the entropy distributions in Fig. \ref{fig:mcd_iou_cost}(a). Comparing 5 MC dropout samples to 20 MC dropout samples shows no significant difference between the distributions of entropies. A t-test between these distributions gives a p-value of 0.6986, allowing us to safely assume these distributions are the same and use 5 MC samples in \gls{ourmethod} for entropy estimation. We account for these 5 MCD samples in all our previous flops calculations.

\section{Conclusion}
\label{sec:conclusion}

In this work, we have developed a computationally parsimonious and more effective alternative to the IDK cascading decision pipeline and demonstrated that our proposed model \gls{ourmethod} outperforms SOTA models on the task of battery material phase segmentation. We also propose a new metric \emph{IoU per GigaFlop} which is useful for characterizing effectiveness of models to yield good predictions at low computational cost. Through various qualitative and quantitative results, we demonstrate that \gls{ourmethod} yields a minimum performance improvement of $\mathbf{174}\%$ on the IoU/GigaFlop metric with respect to compared baselines. We also demonstrate \gls{ourmethod}'s adaptability to complementary models trained on the noisy MNIST dataset, where it outperforms all baselines on IoU/GigaFlop by a miniumum $\mathbf{13.4}\%$. In the future, we shall extend \gls{ourmethod} to incorporate other sophisticated cost metrics and test it in the context of multi-model pipelines comprised of data-driven and scientific simulation models.

\section*{Acknowledgements}
This manuscript has been authored by UT-Battelle, LLC, under contract DE-AC05-00OR22725 with the US Department of Energy (DOE). The US government retains and the publisher, by accepting the article for publication, acknowledges that the US government retains a nonexclusive, paid-up, irrevocable, worldwide license to publish or reproduce the published form of this manuscript, or allow others to do so, for US government purposes. DOE will provide public access to these results of federally sponsored research in accordance with the DOE Public Access Plan (https://www.energy.gov/doe-public-access-plan).

\bibliography{bibfile}

\renewcommand{\appendixpagename}{\centering Appendix}
\appendixpage
\appendix

\section{A: Results \& Discussion}

\begin{figure}[t]
    \centering
    \includegraphics[width=0.4\textwidth]{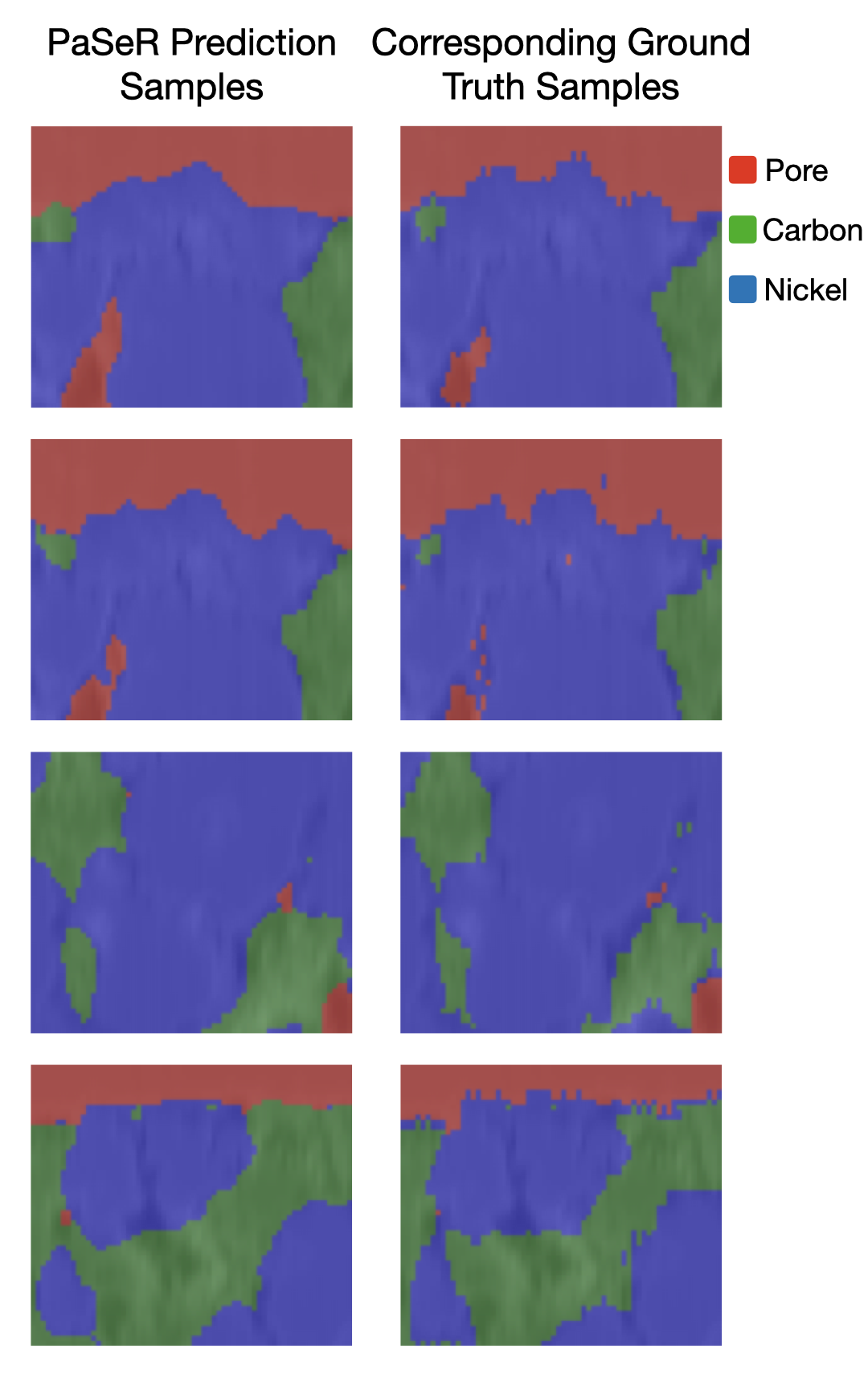}
    \caption{Side by side comparison of \gls{ourmethod} segmentation predictions vs corresponding ground truth. We notice that \gls{ourmethod} yields reliable segmentation results for all classes i.e., pore, nickel, carbon.}
    \label{fig:qualitative_segmentation}
\end{figure}

\begin{figure*}
  \centering
  \includegraphics[width=0.9\textwidth]{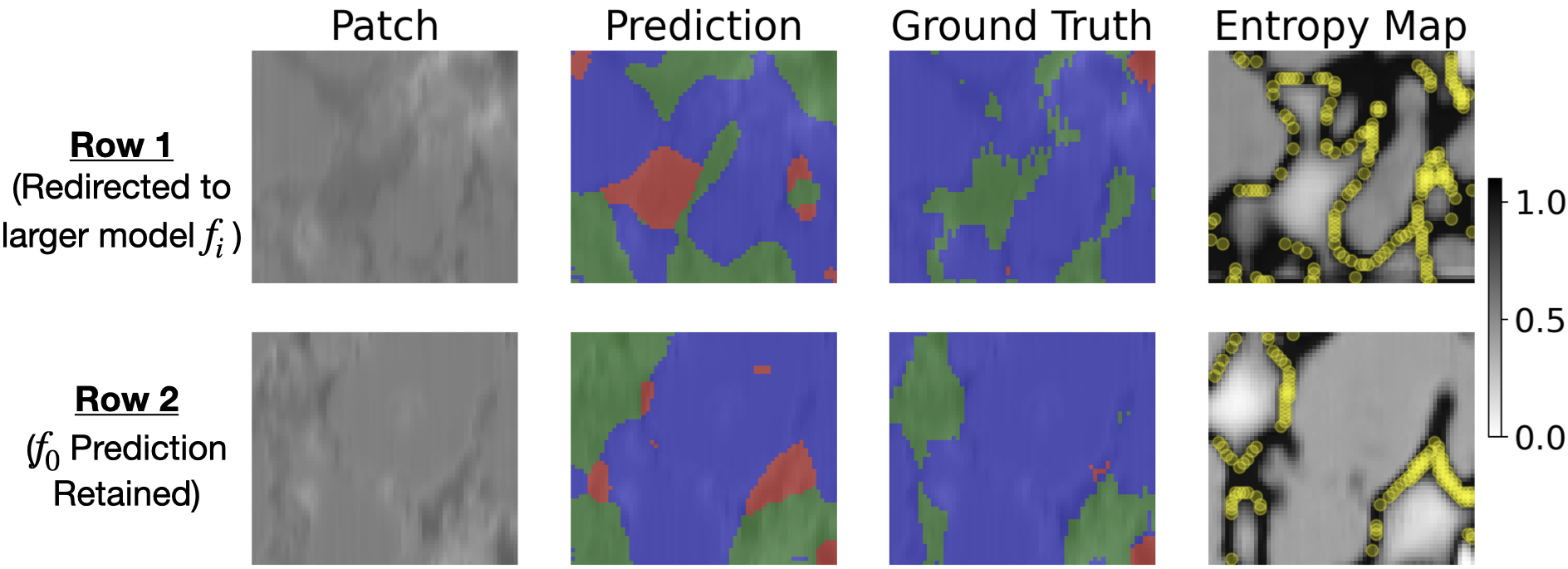}
  \caption{Top row (column 1) depicts an example (low-res) input patch that is consistently redirected by the RL policy ($f_{\mathrm{RL}}$) to higher level models while bottom row (column 1) shows a different patch on which predictions by $f_0$ are retained and no higher-level predictions are solicited by $f_{\mathrm{RL}}$. Column 4 (last column) depicts corresponding entropy maps output by $f_0$ for each patch, while columns 2, 3 depict the predicted segmentation by $f_0$ and the corresponding ground truth segmentation. Yellow pixels on each entropy map depict the pixels wherein $f_0$ yielded entropy greater than a particular threshold $\gamma$. The top row entropy map has \textbf{56}\% more yellow points than the bottom entropy map indicating significantly higher uncertainty of $f_0$ prediction on top image.}
  \label{fig:ent-patches}
\end{figure*}

\subsection{A1: Qualitative Battery Segmentation Results}

In Fig.~\ref{fig:qualitative_segmentation}, we showcase examples of \gls{ourmethod} segmentation performance on the test set ($\mathcal{D}_{\mathrm{test}}$). We notice that \gls{ourmethod} yields good segmentation performance even for the challenging (minority) pore, carbon classes as well as the (majority) nickel class.

We further investigate the performance of the proposed \gls{ourmethod} model by investigating how the predictions and entropy maps of $f_0$ affect the decisions of $f_{\mathrm{RL}}$. Specifically, we show two separate patches in Fig.~\ref{fig:ent-patches} such that the patch in the first row is redirected by $f_{\mathrm{RL}}$ to larger models (i.e., in favor of $f_0$ predictions) while the $f_0$ model predictions are retained by $f_{\mathrm{RL}}$ for the patch in the second row. Although investigating just the low-res input patch (i.e., column 1) of each row, might not yield much insight, the corresponding entropy maps (column 4) showcases that $f_0$ predictions on the patch in the first row are significantly less confident as indicated by the presence significantly larger proportion of yellow points which indicate highest entropy regions (i.e., entropy greater than a pre-set threshold $\gamma$) in a much larger portion of the image patch (relative to patch on the bottom row). Specifically, there are  \textbf{56}\% more yellow pixels in the top row than the bottom row.  The reasoning for such high entropy on the patch in row 1 (relative to row 2) may be gleaned from inspecting the corresponding ground truth images (column 3). We see that the ground truth image in row 1 has significantly more interspersed material phases e.g., the carbon phase - (green) is more interspersed with the nickel (blue) in row 1 than  row 2. On the contrary in row 2, the ground truth depicts a more \emph{segregated} distribution of materials i.e., there exist large contiguous regions of a single material (e.g., large contiguous green, blue regions) which is an easier context for the simpler $f_0$ model to segment (owing to its computational simplicity) relative to its more sophisticated counterparts in the \gls{ourmethod} pipeline. This result further reinforces that \gls{ourmethod} and the RL policy therein (in conjunction with $f_0$) redirect the harder instances (conditioned upon $f_0$ predictions, entropy map and governed by the overall computational cost) to upstream models in a computationally parsimonious manner.

\begin{figure}
    \centering
    \includegraphics[width=0.35\textwidth]{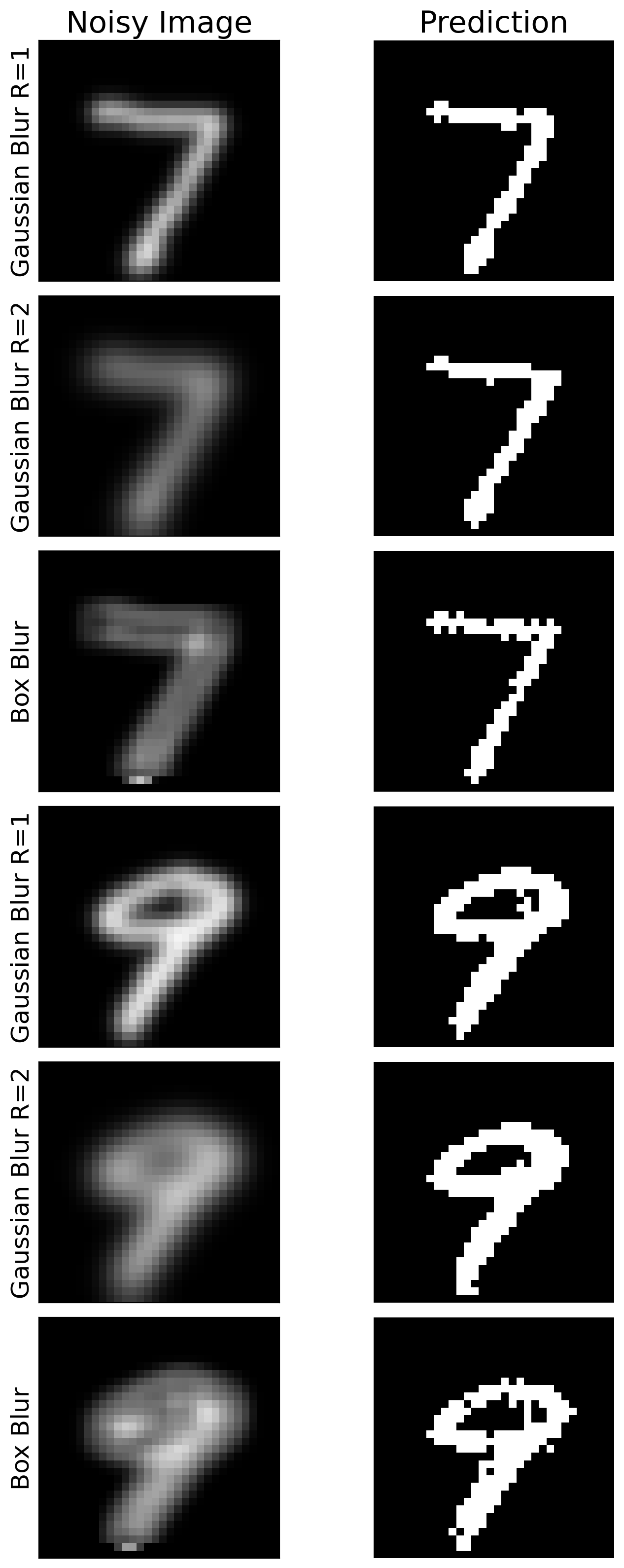}
    \caption{Side by side comparison of Noisy MNIST input images and \gls{ourmethod} predictions.}
    \label{fig:mnist_qualitative}
\end{figure}

\subsection{A2: Qualitative Noisy MNIST Results}

In Fig. \ref{fig:mnist_qualitative} we show noisy MNIST input samples from our test set in the left column and the corresponding foreground/background \gls{ourmethod} segmentation predictions in the right column. Note that the box blur images are the most difficult to de-noise, but \gls{ourmethod} does well on these images because it redirects them to the most sophisticated task model $f_2$.

\section{B: Experimental Setup}

\subsection{B1: Battery Material Phase Segmentation Data}

Our primary dataset consists of 1330 battery phase segmentation tomographic images split into 1270 for training, 20 for validation ($\mathcal{D}_{\mathrm{val}}$), and 40 for testing ($\mathcal{D}_{\mathrm{test}}$). The height and width of each image is 224 x 256. In addition to these images, this dataset contains pixel level annotations of 3 classes: carbon, nickel and pore. We further split the training data into 3 subsets: one for pretraining the segmentation models $\mathcal{D}_{PT}$ with 436 images, another for RL training $\mathcal{D}_{RL}$ with 422 images, and the last for fine-tuning $\mathcal{D}_{FT}$ with 422 images.

\begin{figure}
    \centering
    \includegraphics[width=0.4\textwidth]{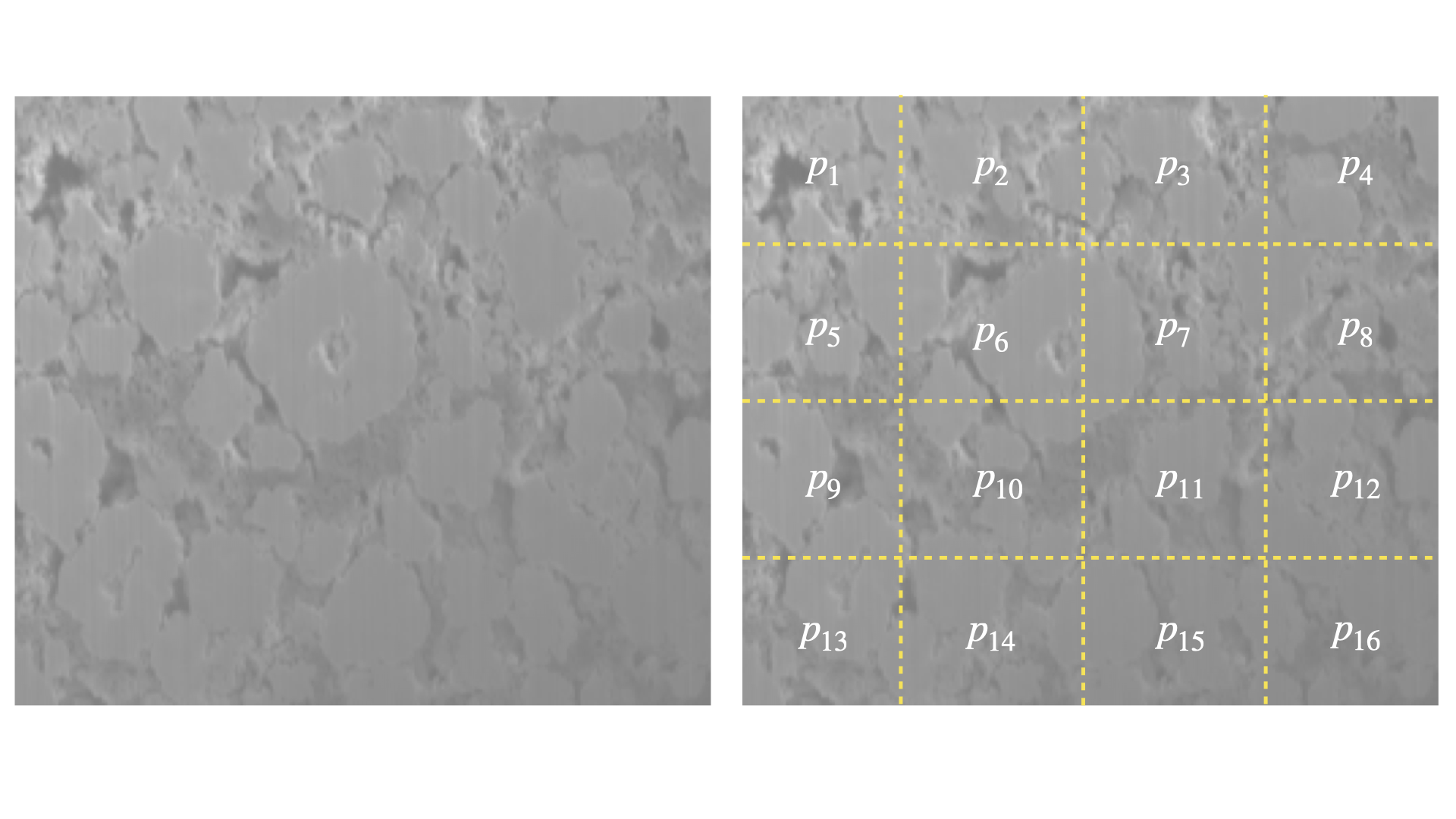}
    \caption{Illustration of our patch splitting method with the full image on the left and the patches shown on the right.}
    \label{fig:image_patches}
\end{figure}

For the input to larger segmentation models $\{ f_1, \ldots, f_m \}$, we split the each image into 16 equal size patches, each of size $56 \times  64$ as shown in Fig. \ref{fig:image_patches}.

\begin{figure}
  \begin{center}
\includegraphics[width=0.4\textwidth]{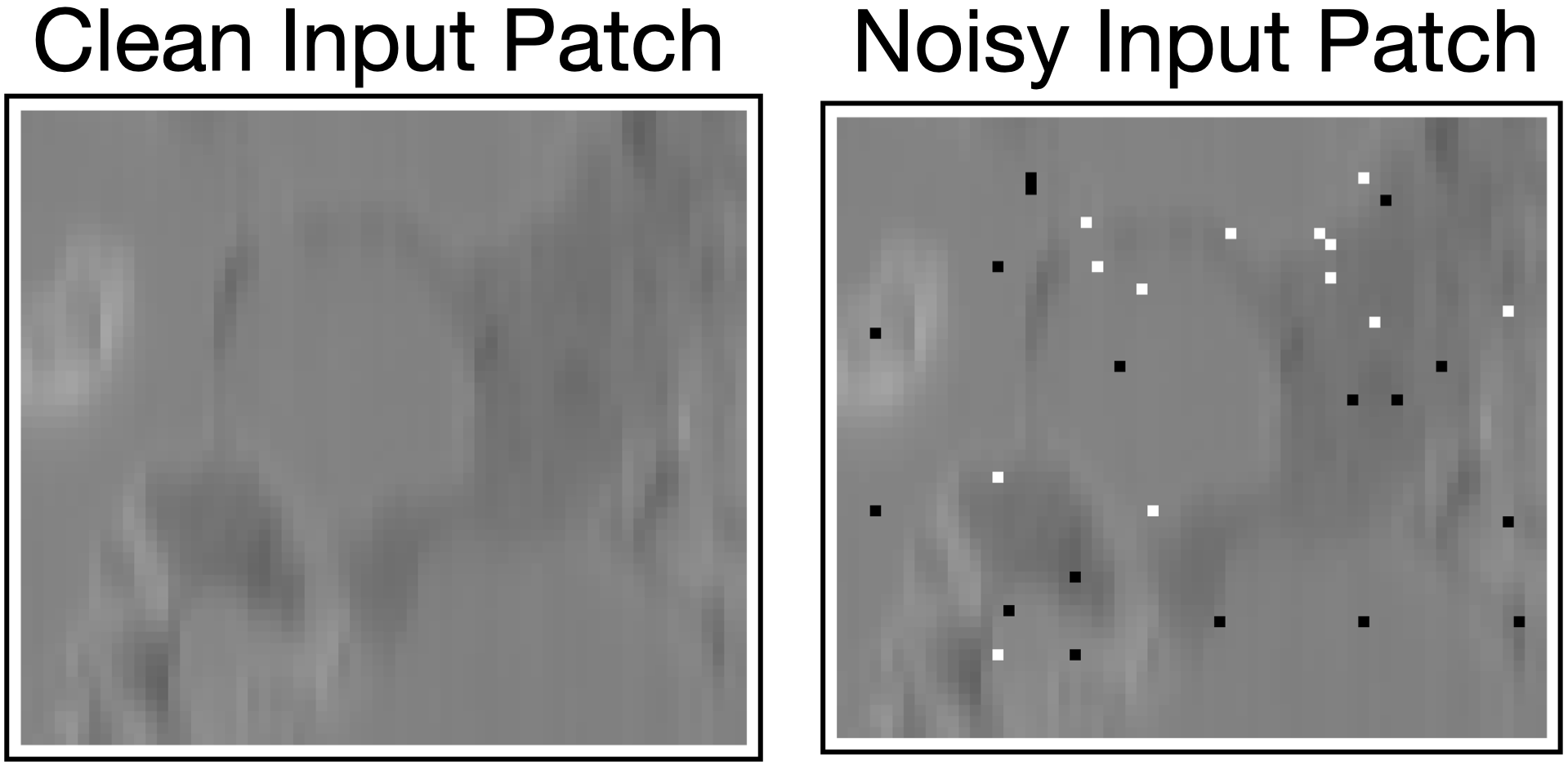}
  \end{center}
  \caption{To test robustness of \gls{ourmethod} pipeline in unseen scenarios (e.g., degradation of input images due to process noise), we add noise to our input data and a sample of one such image patch is depicted above.}
  \label{fig:clean_noisy_patch_image}
\end{figure}

\begin{figure}
  \centering
  \includegraphics[width=0.4\textwidth]{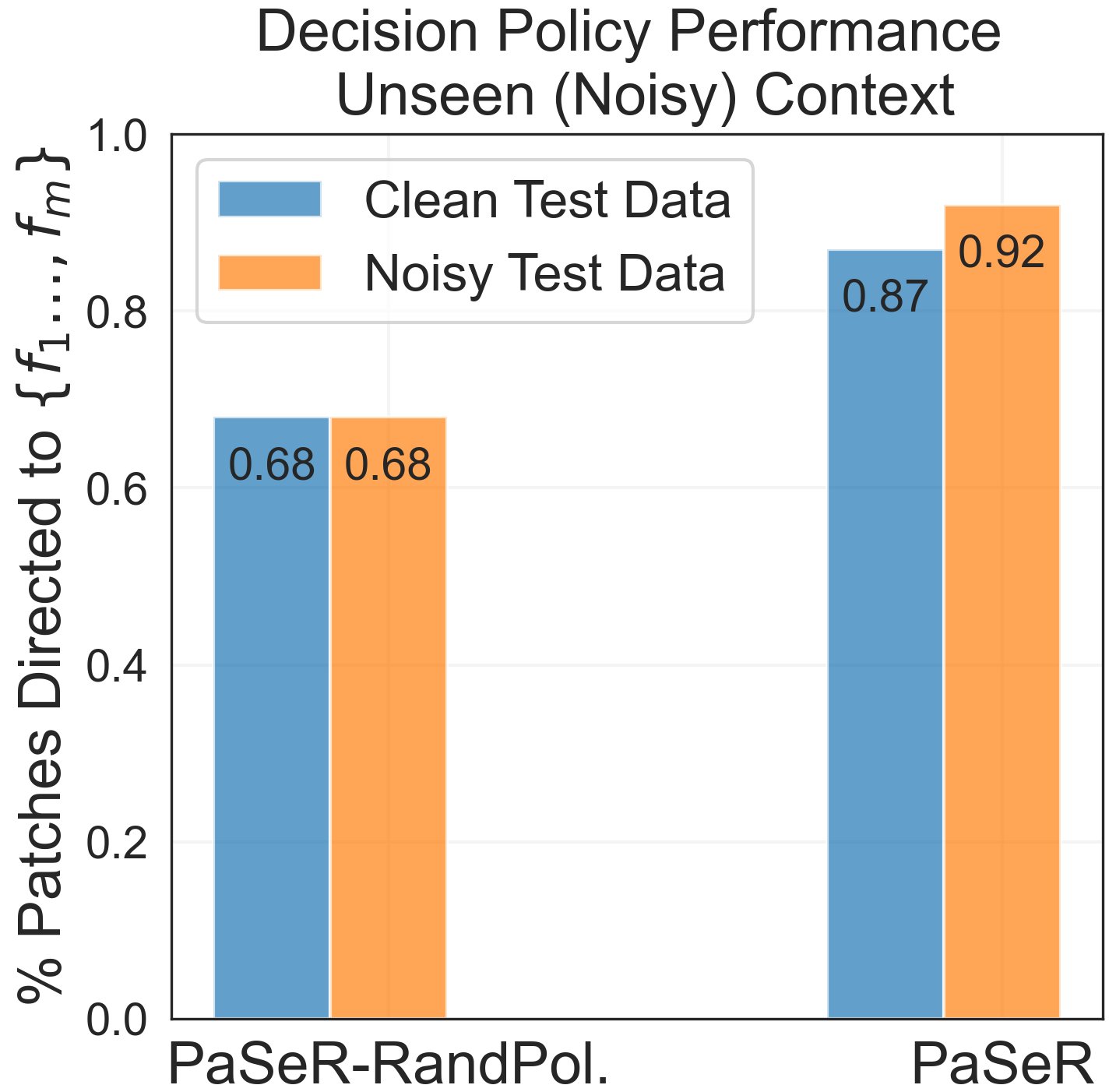}
  \caption{Each bar indicates the percentage of patches sent to larger (i.e. medium or large) models by \gls{ourmethod} vs \gls{ourmethodrandpolicy} when tested on clean (blue) vs (unseen) noisy data (orange). \gls{ourmethod} sends $\sim5.7\%$ more patches to larger models in the unseen (noisy) data case, while \gls{ourmethodrandpolicy} doesn't adapt and sends the same rate ($\sim 68\%$) of patches to the larger models in both cases.}
  \label{fig:percent_patches_redirected_clean_vs_noisy_ours_vs_random}
\end{figure}

\subsubsection{B1.1: Salt \& Pepper Noise}

To investigate the adaptability of our RL policy based \gls{ourmethod} and demonstrate its effectiveness relative to the random policy employed in \gls{ourmethodrandpolicy}, 
we  create a variant of our segmentation dataset injected with salt and pepper noise. This noise was added to every image at a rate of 1\% (i.e., noise is injected into 1\% of the input pixels). For an example of both clean and noisy images see Fig. \ref{fig:clean_noisy_patch_image}.

To highlight the response of the RL policy ($f_{\mathrm{RL}}$), when exposed to this unseen noisy scenario during inference, we capture the number of instances redirected to the medium and large models. Specifically, we capture the percentage of patches in which predictions by $f_0$ were considered under-confident / under-par by $f_{\mathrm{RL}}$ and instead, predictions from more sophisticated models were sought. In Fig.~\ref{fig:percent_patches_redirected_clean_vs_noisy_ours_vs_random}, we showcase this percentage for \gls{ourmethod} and \gls{ourmethodrandpolicy} in the clean data (blue) and noisy data (orange) contexts and notice that despite \gls{ourmethod} RL policy never having encountered noisy data, it is able to recognize that the small model $f_0$ is not confident on unseen instances. Hence, in the noisy scenario, it is able to re-direct a higher percentage of patches to more sophisticated upstream models demonstrating robustness of the RL policy learned by \gls{ourmethod}. In contrast, as \gls{ourmethodrandpolicy} lacks a learnable policy like \gls{ourmethod}, it fails to adapt and sends the same percentage of patches to more sophisticated models in clean and noisy scenarios leading to significantly higher performance degradation.

\subsection{B2: \texorpdfstring{\gls{ourmethod}}{PaSeR} for Battery Material Phase Segmentation Hyperparameters and Model Tuning}

The \gls{ourmethod} model requires a few parameters to be specified. Below we describe each parameter as well as the procedure we used to select their values.

\par
\textbf{Batch Size}. We use a batch size of 32 images, which maximizes the usage of the available GPU memory for the large UNet model. For uniformity, we maintain the same batch size for all models during training. However it must be noted that the batch size for each model can be set to different values. During inference or testing, single images (or batches) may be evaluated.

\par
\textbf{Number of epochs}. The 3 stages of our \gls{ourmethod} model training pipeline: (a) pretraining of segmentation models ($f_0$, $f_1$, $f_2$); (b) training the RL policy ($f_{RL}$); and (c) joint fine-tuning of the segmentation models and RL policy; are executed over the Material Phase Segmentation training datasets \citep{tabassum2022matphase} for 200 epochs. This number of epochs was selected as loss convergence was observed (on a validation set) by this time.

\par
\textbf{Performance-Cost Tradeoff ($\lambda$)}. As the $\lambda$ parameter is introduced in the form of a convex combination in the RL policy reward, the value of $\lambda$ can be varied in the range [0,1]. We evaluated the effect of $\lambda$ by training the RL policy at values of $\lambda = \{ 0.0, 0.1, 0.3, 0.5, 0.7, 0.9, 1.0 \}$. The $\lambda$ parameter is tuned using grid-search only during the RL pre-training stage (the tuned value is used without further updates during fine-tuning). For our battery phase segmentation experiments, we report IoU and IoU/GigaFlop results with $\lambda = 0.5$ because this provides a even balance between task performance and computational cost.

\par
\textbf{Explore Exploit ($\alpha$)}. To encourage the RL policy to periodically explore new actions, we tune an explore/exploit parameter $\alpha$ which determines a ratio between the number of times the policy exploits the action yielding the maximum expected return ($s_{f_{RL}}$)  and the number of times a random action is chosen ($s_U$).
\begin{equation}
    s =
    \begin{cases} 
      s_{f_{RL}} &\text{ with probability } \alpha \\
      s_{U}      &\text{ with probability } 1 - \alpha
    \end{cases}
\end{equation}
During RL pretraining, we start the value of $\alpha$ at 0.7 and adopt a linear schedule to increase it (per epoch) until it reaches an upper limit of 0.95. During the fine-tuning stage, we begin $\alpha$ at 0.95 and linearly increase it to 1.0.
\par
\textbf{Optimizer and Learning Rate $(\eta)$}. We use the Adam optimizer \cite{kingma2014adam} with a learning rate $\eta = 1e^{-4}$.

\subsection{B3: \texorpdfstring{\gls{ourmethod}}{PaSeR} for Noisy MNIST Segmentation Hyperparameters and Model Tuning}

\begin{figure}
  \centering
  \includegraphics[width=0.35\textwidth]{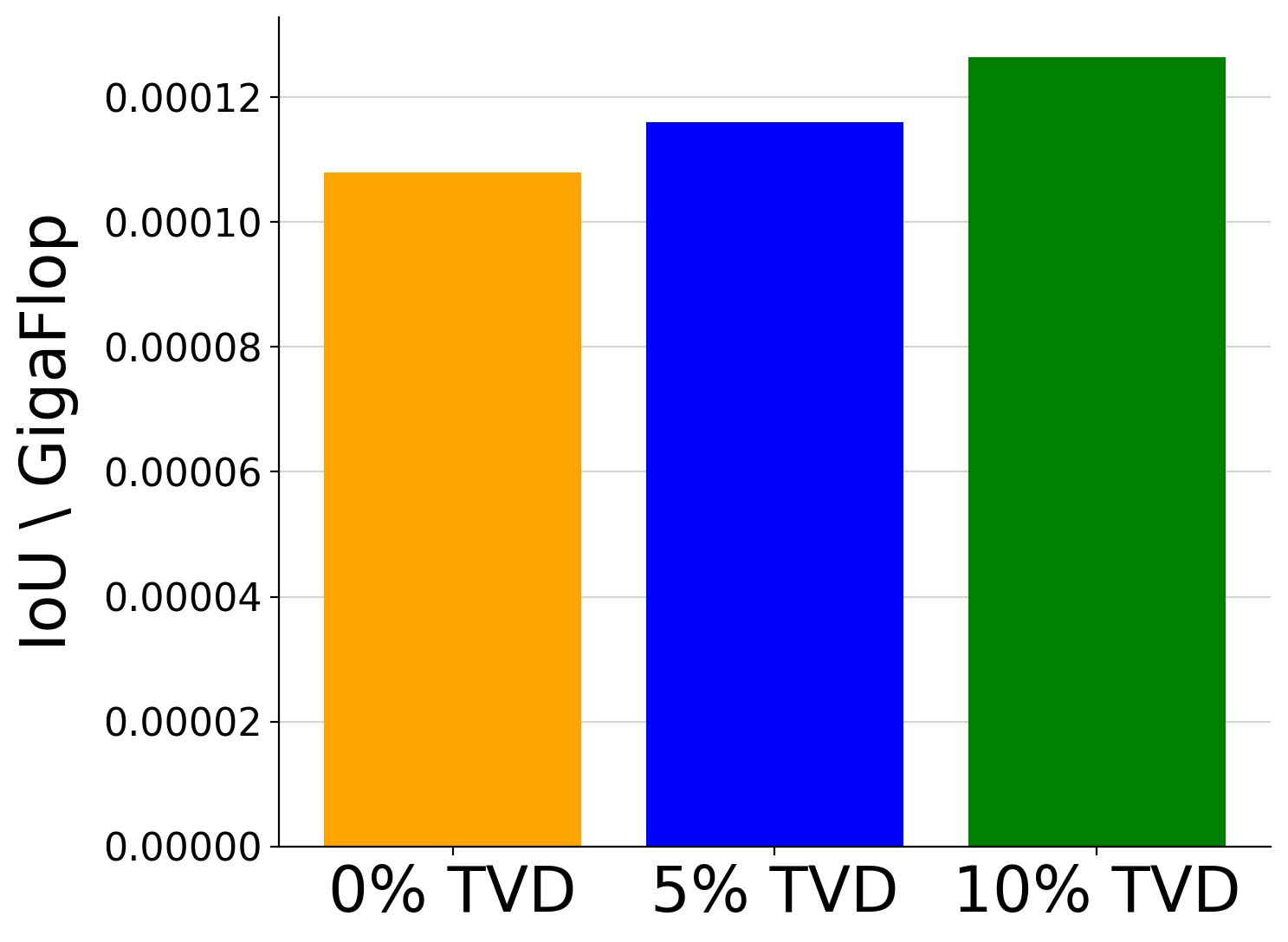}
  \caption{IoU/GigaFlop vs total variation distance (TVD) after fine-tuning on the noisy MNIST dataset. As we increase the TVD threshold, the IoU/GigaFlop increases because \gls{ourmethod} sends a greater proportion of images to the smaller models.}
  \label{fig:iou_tvd_plot}
\end{figure}

We train and run \gls{ourmethod} on the Noisy MNIST dataset with slightly different parameters than the Battery segmentation dataset in order to demonstrate the adaptability of \gls{ourmethod}'s RL policy.

\par
\textbf{Batch Size.} We set batch size to 128 because of the relative small size of each instance in this dataset (32 x 32 images).

\par
\textbf{Performance-Cost Tradeoff ($\lambda$).} During RL pre-training on this dataset, we set $\lambda = 0$ so that the RL policy learned by \gls{ourmethod} is optimal in terms of IoU without regard for computational cost. Once we have learned this optimal model assignment policy from the data, we set a total variation distance (TVD) percentage threshold from this optimal distribution. During fine-tuning, we increase $\lambda$ on a linear schedule until the newly fine-tuned \gls{ourmethod} RL policy reaches this TVD threshold, at which point we stop fine-tuning. By increasing $\lambda$, \gls{ourmethod} is able to trade-off performance for computational cost until the desired deviation from the optimal policy is achieved. In Fig. \ref{fig:iou_tvd_plot}, we plot the effect of increasing TVD thresholds on IoU/GigaFlop efficiency. As expected, increasing the TVD threshold increases the IoU/GigaFlop efficiency of the fine-tuned model. The 0\%, 5\%, and 10\% TVD fine-tuned models have test set IoUs of 0.8432, 0.8328, and 0.8231 respectively.

In real-world applications of \gls{ourmethod}, the TVD percentage threshold is a straightforward way for domain experts to directly trade-off cost and performance in a dynamic fashion. Consider the case of quality control (QC) for battery manufacturing. In some scenarios, such as small battery manufacturing (AA batteries for example) we would choose a high TVD threshold because the cost of each battery is low and our goal is to quickly and efficiently manufacture them rather than ensure the highest possible quality. However in the case of electric vehicle (EV) batteries, we would set a low TVD threshold because we wish to ensure that the quality of each battery is high and to reduce the chance of early failure or degradation.

\subsection{B4: \texorpdfstring{\gls{idkbaseline}}{IDK-Cascade} Hyperparameters and Model Tuning}

For the \gls{idkbaseline} model, we setup a cascade with the same three segmentation models as \gls{ourmethod} (small, medium and large UNets). For each segmentation model $f_0, f_1$, the \gls{idkbaseline} model uses an entropy threshold to decide if a patch should be passed to the next larger model. In addition to these thresholds, we use the same cost function as \gls{ourmethod} with a cost parameter $\lambda_{IDK}$ weight where $\mathcal{L}(\hat{\by}_i, \by)$ is the cross entropy loss:
\begin{equation}
    \label{eq:loss_idk}
    l_{IDK} = \mathcal{L}(\hat{\by}_i, \by) + \lambda_{IDK} \cdot C(f_i)
\end{equation}
To select the optimal values for entropy thresholds $\alpha_{f_i}$, we measure the distribution of entropy values in the validation dataset ($\mathcal{D}_{\mathrm{val}}$) for each model and do a grid search between one standard deviation below and above the mean.

\par
\textbf{Small UNet Entropy Threshold ($\alpha_{f_0}$)} For the small UNet entropy threshold, we grid search between $[0.61, 0.72]$. This value range was selected as it spanned one standard deviation (above and below) away from the mean entropy computed on the validation set ($\mathcal{D}_{\mathrm{val}}$).
\par
\textbf{Medium UNet Entropy Threshold ($\alpha_{f_1}$)} For the medium UNet entropy threshold, we grid search between $[0.20, 0.35]$ This value range was selected as it spanned one standard deviation (above and below) away from the mean entropy computed on the validation set ($\mathcal{D}_{\mathrm{val}}$).
\par
\textbf{Cost weight ($\lambda_{\mathrm{IDK}}$)} In \cite{wang2017idk}, the authors use $\lambda_{IDK} = 0.04$. We follow their example and grid search between $[0.0, 1.0]$ and find that $\lambda_{IDK} = 0.01$ minimizes the loss in Equation \ref{eq:loss_idk}

\end{document}